\documentclass{article}

\usepackage[utf8]{inputenc}
\usepackage[T1]{fontenc}
\usepackage{babel}
\usepackage{amsfonts}
\usepackage{geometry}
\usepackage{array}
\usepackage{multirow}
\usepackage{amsmath}
\usepackage{amssymb}
\usepackage{graphicx}
\usepackage{setspace}
\usepackage{esint}
\usepackage{rotating}
\usepackage{color}
\usepackage{eurosym}
\usepackage{comment}
\usepackage{longtable}
\usepackage{subcaption}
\usepackage{float}
\usepackage{titlesec}
\usepackage{booktabs}
\usepackage[table]{xcolor} 

\title{Driver Assistance System Based on Multimodal Data Hazard Detection}

\begin{document}

\begin{titlepage}
    \centering
    \vspace*{3cm}
    
    {\Huge\bfseries\fontfamily{ptm}\selectfont Driver Assistance System Based on Multimodal Data Hazard Detection \par}

    \vspace{3cm}
    {\Large Long Zhouxiang, Ovanes Petrosian\par}
    
    \vspace{2cm}
    {\large \today\par}
\end{titlepage}

\section{Introduction}
Advances in artificial intelligence have given a significant boost to autonomous driving technology. However, the advantages of deep learning methods for the control process of autonomous driving have not been thoroughly studied. For example, in the United States alone, an average of 6 million car accidents occur each year, in which about 3 million people are injured and about 2 million suffer permanent injuries\textsuperscript{\cite{ aydelotte2017crash}}. Therefore, the proposed automatic driving hazard detection technology can not only improve the safety of driving, but also can greatly reduce the workload of drivers. Currently, automatic driving technology is mainly researched from road detection, expression recognition, and speech recognition.

The challenging task of developing an automated driving system involves sensing and detecting the road so that the driver can receive immediate alerts of any potential dangers and risks. The authors in \textsuperscript{\cite{xiao2018hybrid}} proposed a novel hybrid CRF model for road detection by considering contextual correlations between different modalities. RBNet\textsuperscript{\cite{chen2017rbnet}} learned an advanced CNN feature that allows one-step detection of road regions and their edges. In\textsuperscript{\cite{teichmann2018multinet}}, the authors proposed a joint encoder-decoder scheme which has higher accuracy. \textsuperscript{\cite{bayoudh2021transfer}} combines pre-training parameters from the EfficientNet-B0 baseline \textsuperscript{\cite{tan2019efficientnet}} and a shallow 3D CNN to maximize the overall performance of road segmentation.

On the other hand, it is to detect the driver's state. For example, the driver's expression or drowsiness level.Lyu et al.\textsuperscript{\cite{lyu2018long}} proposed a deep framework based on multi-granularity by intelligently using CNNs and LSTMs for drowsiness detection in videos. Vijayan and Sherly\textsuperscript{\cite{ vijayan2019real}} proposed three CNN architectures including ResNet50, VGG16, and InceptionV3 for sleepiness detection in first-person driver videos. These models are trained by fusing them together using a feature fusion architecture layer. A similar approach was taken by Park et al.\textsuperscript{\cite{ park2016driver}}, who integrated the results obtained by AlexNet, VGG-FaceNet and FlowNet through a fully connected layer for sleepiness detection. Facial expressions include happiness, surprise, anger, sadness, fear, and disgust\textsuperscript{\cite{priestley1999expression}},and these responses are important for driving vehicle safety. Driver emotions should be able to support driving abilities such as attention, good judgment, sound decision making, and fast response time\textsuperscript{\cite{ eyben2010emotion}}.
Agrawal et al.\textsuperscript{\cite{agrawal2013emotion}} uses a fuzzy rule-based system (FBS) to handle simultaneous facial gesture tracking and emotion recognition, and the car switches to automatic mode when any sign of driver inattention or driver fatigue.  Goutam et al. \textsuperscript{\cite{sahoo2023performance}} used the parameters of AlexNet, SqueezeNet and VGG19 models through transfer learning to develop a facial expression recognition system that will monitor the driver's facial expressions to recognize their emotions and provide instant help for safety.

The aim of this paper is to develop a multimodal driver assistance detection system that conforms road condition video data with expression video data and audio data

Advanced driver assistance systems (ADAS), a frontier autonomous driving technology, have immense potential to revolutionize transportation.However, accurately perceiving and identifying driving anomalies is a major challenge. Driving scenarios follow a long-tailed distribution, with normal conditions dominating, while a vast array of anomalous events (e.g., sudden lane changes, unexpected obstacles, driver fatigue/illness) constitute the minority. Although rare, handling these anomalies is crucial for autonomous systems\textsuperscript{\cite{9111005}}. 
\\
Current research on detecting driving incidents largely relies solely on the single modality of road condition video data recorded by dashboard camera\textsuperscript{\cite{9022086}}. These studies mainly focus on identifying abnormal frames in the video data. However, due to the long-tailed distribution of driving events, existing training data struggles to cover all rare driving scenarios, and the immense uncertainty of the real world also implies an inability to predict all possible occurrences\textsuperscript{\cite{7736125}}. In fact, some studies indicate that in the field of autonomous driving, vehicles need to undergo tens of billions of miles of testing to generate sufficient data on rare occurrences to meet 
training\textsuperscript{\cite{KALRA2016182}}.
\\
Relying solely on single-modality data for judgment, if errors occur, will lead to extremely severe consequences. Moreover, considering road safety and relevant laws and regulations, no country has yet achieved true driverless driving, with current autonomous driving systems still relying on the driver as the primary entity, serving an assistive role. Therefore, incorporating other driving information such as driver video and audio for auxiliary judgment not only allows for more accurate identification of various incidents but also aligns with the current practical application scenario demands. Even for some extremely rare driving incidents, although training data cannot fully simulate them, relying on human drivers' experience and common sense can still enable rapid reactions, providing assistance for judging the current driving status\textsuperscript{\cite{7972192}}.\\
Based on the above considerations, this paper proposes a multimodal data input assisted driving model to determine the current driving state. The model simultaneously utilizes road condition video data, driver facial video data, and driver audio data as inputs to avoid potential misjudgments caused by a single data source and mitigate the impact of the long-tailed distribution of driving events.
\\
Multimodal methods are a class of techniques that operate jointly on multiple data types. In our model, we primarily investigate data composed of an audio modality and two visual modalities, employing data fusion techniques to consider the interplay between different modalities. Early fusion may not be suitable for extremely dissimilar data types, while late fusion mainly considers the independently learned features of each modality. For audio-visual multimodal data, intermediate feature fusion could potentially yield better performance\textsuperscript{\cite{9956592}}.
\\
In recent transformer-based architectures for multimodal data recognition tasks, most research has focused on utilizing pre-extracted features, which are subsequently fused with the learning model, rather than creating end-to-end trainable models\textsuperscript{\cite{Tsai2019MultimodalTF}} \textsuperscript{\cite{N2020MultimodalER}}. This limits the applicability of such methods in real-world scenarios, especially in assisted driving scenarios where feature extraction poses significant challenges and introduces uncertainty into the entire processing pipeline. This issue is particularly pronounced for methods using audio information, as audio signal records are rarely available in practical applications and need to be estimated separately\textsuperscript{\cite{Tsai2019MultimodalTF}} \textsuperscript{\cite{N2020MultimodalER}}. Therefore, our work constructs an end-to-end model (without the need for pre-learned features) and performs fusion at the intermediate level, demonstrating the model's robustness to incomplete or noisy data. samples\textsuperscript{\cite{Bondi:2018:AWS:3209811.3209880}}.
\begin{figure}[htbp]
  \centering  \includegraphics[width=0.7\textwidth]{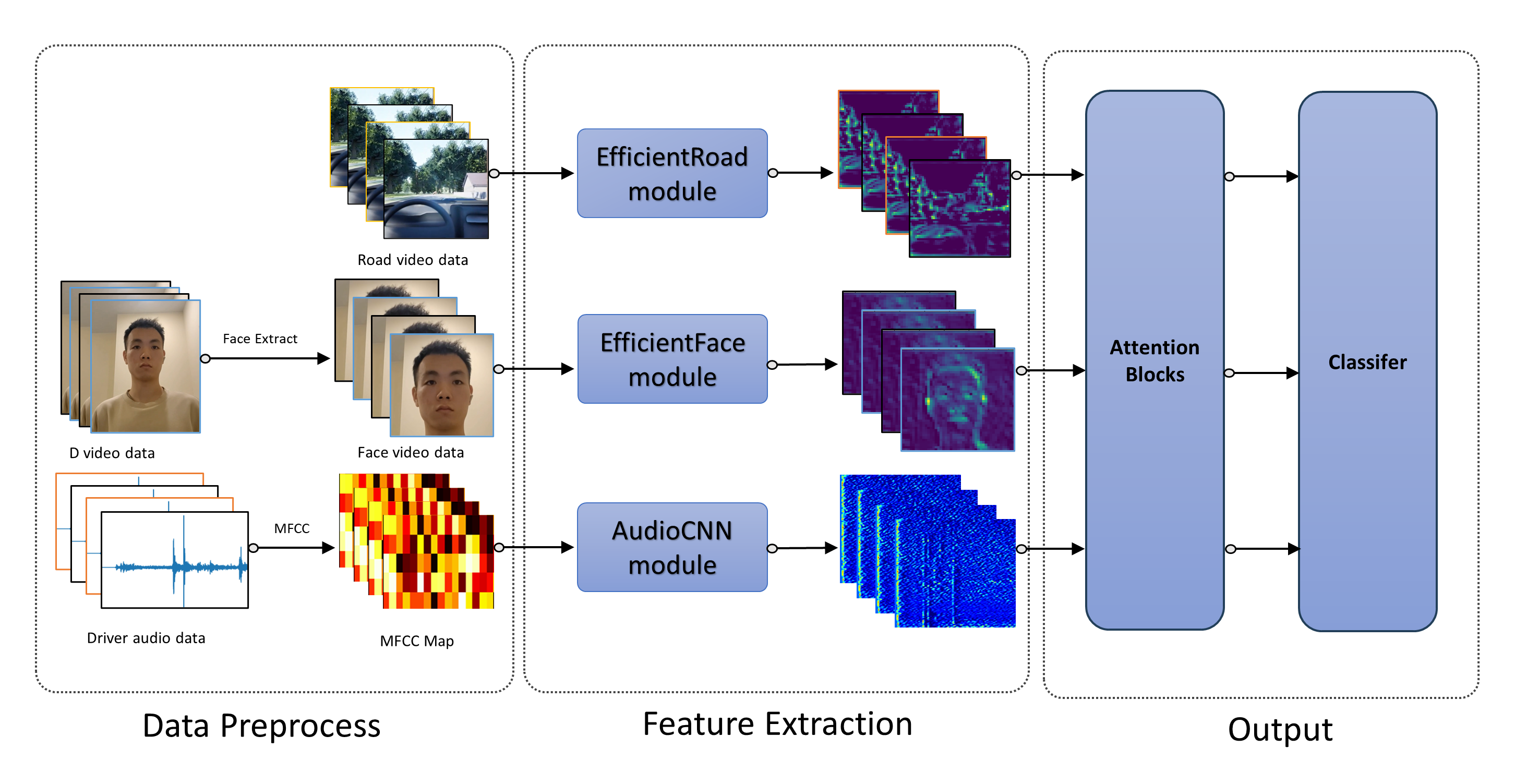}
  \caption{End-to-end recognitions framework}
  \label{fig:end}
\end{figure}
\\
Currently, there is a lack of publicly available datasets containing road condition video data, driver facial video data, and driver audio data simultaneously. Consequently, this paper establishes a three-modality dataset by using the Arisim driving simulation project and Logitech driving simulation devices to conduct simulated driving while recording road condition video data, driver facial video data, and driver audio data during the simulated driving process, thereby generating the required data.\\
The main contributions of this paper are as follows:
\sloppy
\begin{itemize}
\item We propose an innovative multimodal fusion framework that incorporates road condition video data, driver video data, and driver audio data as three input modalities, achieving end-to-end learning from raw data without relying on separate feature extraction.
\item We adopt an attention-based intermediate fusion strategy, pairwise fusing different modalities, which effectively captures cross-modal correlations and improves the utilization efficiency of multimodal information.
\item We simulate the driving process and generate an annotated dataset containing three modalities, providing a valuable data resource for multimodal driving behavior analysis.
\end{itemize}
\section{Related work}
Currently, the most objective and commonly used VAD datasets are mainly sourced from various surveillance cameras. For example, the UCSD\cite{data1} and ShanghaiTech \cite{data2} datasets have recorded anomalous situations under different lighting conditions and scenes through campus surveillance cameras. The UCF-Crime \cite{data3} dataset focuses on public safety, including anomalous events such as traffic accidents, burglaries, and explosions.

With the development of assisted driving technology, anomaly detection in first-person perspective traffic videos has attracted widespread attention. Chan et al.\cite{data4} proposed a dataset of road traffic accident videos recorded by dash cameras, with anomalous frames annotated. {data5} collected 1,500 video clips of road anomalies recorded by dash cameras from East Asia on YouTube, with the start and end times of anomalous events labeled. Additionally, \cite{9022086} extracted 803 clips from the BDD100K\cite{data6} dataset to construct a new dataset.

In the realm of driver video detection, L. Lyu et al. \cite{lyu2018long}presented the NTHU-DDD dataset, which is designed to detect driver drowsiness. This dataset captures a variety of driver behaviors under different lighting conditions, including normal driving, yawning, slow blinking, and dozing off.The KMU-FED dataset \cite{sahoo2023performance} captures the facial expressions of drivers using an infrared camera while driving, including 55 image sequences from 12 subjects.Although there is currently a lack of publicly available annotated datasets specifically targeting facial expressions and speech anomalies of drivers, there are many valuable datasets in the field of human video and audio emotion classification research that can serve as references. For example, video datasets annotated based on discrete emotion classification, such as RAVDESS and RAV-DB\cite{RAVDESS}, provide valuable resources for research. Additionally, some datasets employ more complex emotion classification methods, such as valence and arousal-based emotion classification, like AffectNet\cite{AffectNet} and RECOLA\cite{RECOLA}. These datasets offer important references and insights for exploring anomaly detection in drivers' emotional states.

Extensive research has been devoted to addressing road incident detection and multimodal recognition.These efforts contribute significantly to the development of intelligent transportation systems by leveraging advanced sensor technologies and sophisticated algorithms to promptly identify potential hazards on the road.

The EfficientFace framework used in video branch processing is a highly efficient network structure that balances computational efficiency and recognition accuracy. It utilizes EfficientNet-B5 as its backbone network and employs depthwise separable convolutions to reduce the number of parameters while maintaining a high level of feature extraction capability. Additionally, it incorporates residual connections to facilitate information flow and gradient propagation within the network.

Furthermore, EfficientFace integrates an attention mechanism module, which enhances its ability to detect occluded parts in images and recognize key regions. The aim of EfficientFace is to provide a compact and efficient image detection framework by optimizing the sharing and learning of features of different scales and modes within the network structure\cite{Wang2023EfficientFace}.

Mel-Frequency Cepstral Coefficients (MFCC) are a commonly used acoustic feature in tasks like speech recognition, speaker identification, and other speech-related tasks. MFCC is used to extract features from speech data by first taking the logarithm of the speech spectrum on the Mel frequency scale, and then applying a discrete cosine transform to the processed results to obtain the cepstral coefficients\cite{MFCC}.

% The process of extracting MFCC is as follows:

% \begin{enumerate}
%     \item \textbf{Pre-emphasis}: Apply a high-pass filter to the speech signal to enhance the high-frequency parts of the speech data.
%     \item \textbf{Framing}: Divide the speech signal into short frames, each with a length of about 20 to 30 milliseconds.
%     \item \textbf{Windowing}: Apply a Hamming window to each speech segment that was framed in the previous step.
%     \item \textbf{Short-Time Fourier Transform (STFT)}: Perform a Fast Fourier Transform (FFT) on the signal data processed by windowing to obtain the spectrum.
%     \item \textbf{Mel-scale Filterbank}: Pass the transformed spectrum data through a set of triangular filters to obtain the Mel spectrum.
% \end{enumerate}

% MFCC is an auditory feature that fully considers the hearing characteristics of the human ear, making the data processing more aligned with the human auditory perception process. MFCC is widely used in tasks such as speech emotion recognition and speaker recognition. By processing speech data in this way to extract MFCC features, it not only retains the important and key information of the speech signal but also enhances the discriminability of the extracted features, thus performing excellently in various speech processing tasks.

For intermediate feature fusion, the commonly used approach involves sharing features at the intermediate layers of a neural network. This process begins with performing feature extraction on each modality or data source separately. After extracting the features, they are fused to jointly learn the feature representations of different modalities. The fusion methods include concatenation, addition, weighted average, and the attention mechanism, which will be introduced later. Using intermediate feature fusion can retain the feature representations of each modality or data source to a certain extent, thereby preserving their advantages. Additionally, the choice of fusion layers is relatively flexible.\cite{9068523}.

In the field of modality fusion, the use of self-attention mechanisms is currently a relatively effective fusion method\cite{vaswani2017attention}. The formula \( A_n \) represents the output of a self-attention mechanism, which is a commonly used technique in sequence modeling. This formula calculates attention weights by applying the softmax function to the dot product of the query vectors \( q \) and the key vectors \( k \), and then these weights are applied to the value vectors \( v \), resulting in a weighted output.

Specifically, \( qk^T \) captures the similarity between the queries and the keys, computed via the dot product. This is then divided by \( \sqrt{d} \), where \( d \) represents the dimensionality of the latent space, to control the scaling of the variables before the softmax function, aiding in gradient stability. The softmax function is then applied to ensure that the sum of all weights equals 1, representing a probability distribution.

Thus, \( A_n \) reflects a weighted representation of the input features after undergoing a series of learnable transformations, allowing the model to dynamically allocate importance among different parts of the input. This mechanism is a cornerstone of the Transformer architecture, which has demonstrated superior performance across a variety of tasks and applications.

\begin{equation}
A_n = \text{softmax}\left(\frac{qk^T}{\sqrt{d}}\right)v,
\end{equation}
where \( d \) is the dimensionality of a latent space, for the vector \( d \) is the dimensionality of the vector, and for matrix the \( d \) respresents the dimensionality of the column of the matrix.Considering the task of fusion of two modalities \( a \) and \( b \), self-attention can be utilized as a fusion approach by calculating queries from modality \( a \) and keys and values from modality \( b \). This results in representation learnt from modality \( a \) attending to corresponding modality \( b \) and further applying the obtained attention matrix to the representation learnt from modality \( b \).

This process of calculating attention weights and applying them to the value vectors allows the model to focus on different parts of the input sequence with varying degrees of importance, thereby capturing complex dependencies and relationships within the data.And the self-attention mechanism (Self-Attention) has achieved remarkable performance in multimodal fusion in recent years.

% The self-attention mechanism can adaptively capture long-range dependencies between different modalities and dynamically modulate the interactions among modal features, thereby more effectively modeling cross-modal correlations. Specifically, self-attention computes the similarities between Query, Key, and Value to perform a weighted sum of features at different positions, capturing global information. Multi-Head Attention further enhances the model's expressive power by computing attention from different subspaces. The self-attention mechanism addresses the limitations of convolutional neural networks in modeling long-range dependencies, demonstrating outstanding multimodal fusion capabilities.

\newpage
\section{Model}
In this section we wiil present the Tri-modal (driver audio - driver video - road video) recognition framework. As shown in Figure 1, the model is divided into three branches that learn the feature representations of the driver's audio data, the driver's visual data, and the remote sensing visual data of the road information, respectively. We adopt a mid-fusion mechanism to fuse any two modalities' features and jointly learn the fused feature representations. In all three branches, we utilize 1D convolutional blocks to capture temporal information. The detial structure of each branch is shown in Table 1.
\\
Next I will introduce the model from the three modality branches.
\\
\textbf{Face Video:}\\For the driver face video branch of the model, in order to implement an end-to-end trainable model that is capable of learning from raw video, we have incorporated the feature extraction part as a component of our pipeline and optimized it together with the multimodal fusion module. We employed a feature extractor based on the EfficientFace framework, which utilizes 1x1 pointwise convolutions to expand the channel dimension of the input feature maps, followed by depthwise convolutional layers that apply an independent convolutional kernel to each input channel, extracting local features within the channels. The output feature maps are then compressed back through another set of 1x1 pointwise convolutions. Additionally, residual connections within each feature extractor block are used to promote the backward propagation of gradients and alleviate the vanishing gradient problem commonly encountered in deep networks. After extracting features from each frame of the driver face video using the feature extractor, these are fed into subsequent 1D convolutional blocks. As opposed to the traditional approach of processing video data with 3D convolutional blocks, we opt for extracting facial features with the feature extractor and temporal features with 1D convolutional blocks. This method not only reduces computational overhead but is also sufficient for our model, which focuses on real-time assessment of the current driving condition for potential hazards; hence, the sequential information of the entire video is not significantly crucial for our task, and 1D convolutional blocks are adequate for capturing the necessary temporal information.
\\
Although our model is an end-to-end model and the entire system learns and optimizes simultaneously, the facial visual feature extraction part of the model can still be considered as an independent component. The second part of the model only requires the features extracted by the feature extraction part without concerning how the features were extracted. Therefore, in our model, the feature extraction portion can be viewed as a replaceable module. In the second part, we further use convolutional blocks to learn temporal features, each consisting of a 1D convolutional layer with 3×3 kernels, batch normalization, and ReLU activation. Further details can be seen in Table 1, where k denotes the kernel size, d represents the number of filters in the convolutional layers, and s signifies the stride. The convolutional blocks are grouped into two stages, used for the further described multimodal fusion.
\\
\textbf{Road Video:}
\\
For the road video branch of the model, considering the need for data fusion, we employed a structure similar to that of the driver's face video branch to facilitate the fusion of data. Additionally, we perform channel shuffling operations, where the feature maps are divided into several groups along the channel dimension, and then these groups of feature maps are concatenated along the same dimension. This helps facilitate inter-group information exchange and implements attention mechanisms in both the channel and spatial dimensions. By learning to generate weights for the channel and spatial dimensions, the feature maps are adaptively calibrated to emphasize channels and areas with more information.
\\
\textbf{Driver Audio:}
\\
For the driver's audio branch of the model, we similarly employ four convolutional blocks, each comprised of a 1D convolutional layer, batch normalization, a ReLU function, and a max-pooling layer. We utilize Mel-frequency cepstral coefficients (MFCCs) as audio features. Initially, the speech signal is subjected to high-pass filtering, followed by framing and windowing. Subsequently, a fast Fourier transform is applied to each frame to obtain the spectrum. The spectrum is then filtered through a Mel filter bank, and the output is transformed via the discrete cosine transform (DCT) to obtain the MFCCs.
\begin{figure}[htbp]
  \centering  \includegraphics[width=0.7\textwidth]{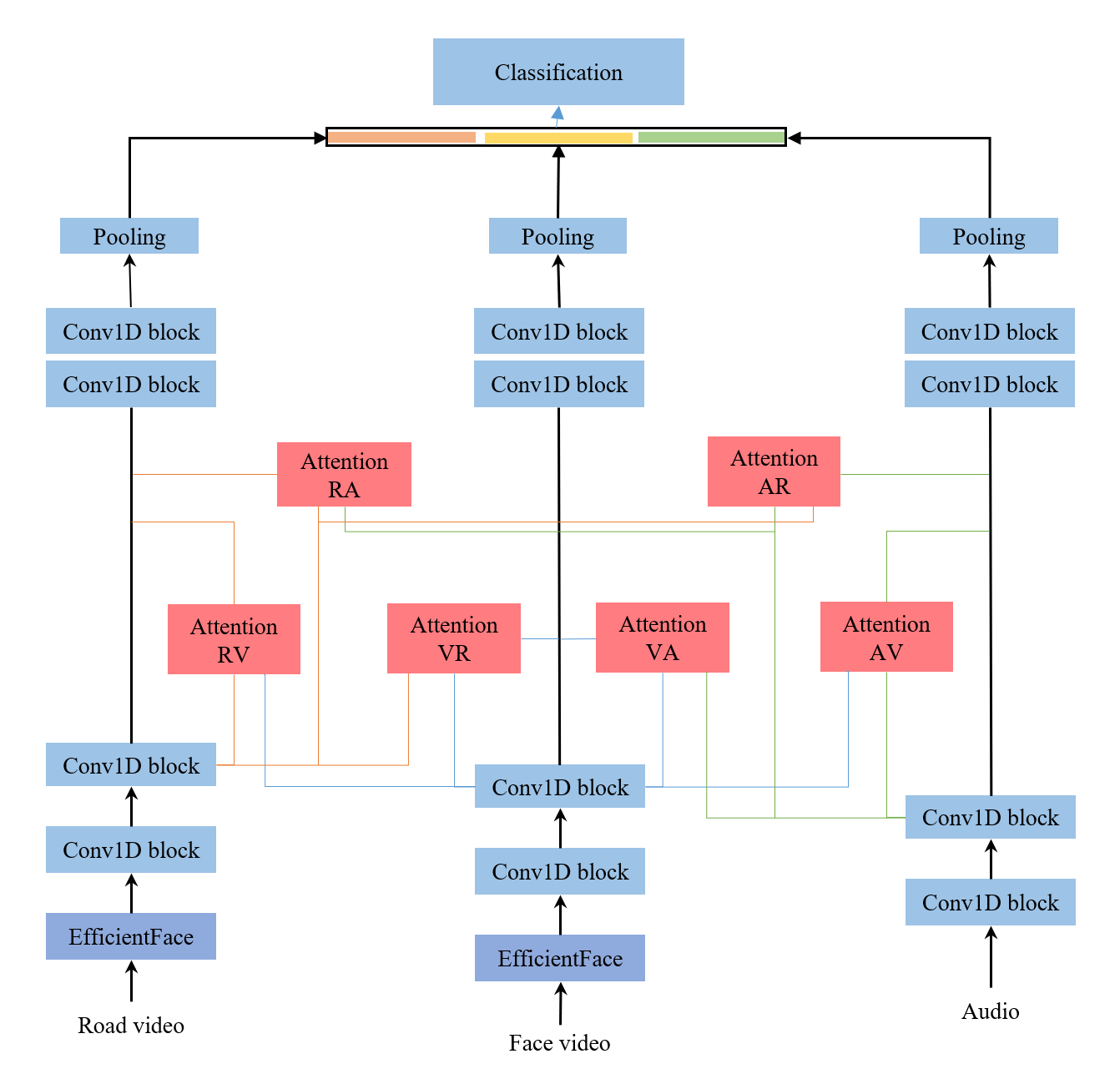}
  \caption{Tri-modal recognitions framework}
  \label{fig:tri-modal recognitions framework}
\end{figure}
\begin{table}[h!]
\centering
\caption{Architecture of the visual and audio modules}
\label{tab:table1}
\begin{tabular}{@{}ll@{}}
\toprule
\multicolumn{2}{c}{\textbf{Architecture of the video(road) branch}} \\
\midrule
Stage1 & \begin{tabular}[c]{@{}l@{}}efficientRoad module\end{tabular} \\
\midrule
Stage2 & \begin{tabular}[c]{@{}l@{}}Conv1D [k=3, d=64, s=1] + BN1D + ReLU\\ Conv1D [k=3, d=64, s=1] + BN1D + ReLU\end{tabular} \\
\midrule
Stage3 & \begin{tabular}[c]{@{}l@{}}Conv1D [k=3, d=128, s=1] + BN1D + ReLU\\ Conv1D [k=3, d=128, s=1] + BN1D + ReLU\end{tabular} \\
\midrule
Predict & Global Average Pooling + Linear \\
\midrule
\multicolumn{2}{c}
{\textbf{Architecture of the video(driver) branch}} \\
\midrule
Stage1 & \begin{tabular}[c]{@{}l@{}}efficientFace module\end{tabular} \\
\midrule
Stage2 & \begin{tabular}[c]{@{}l@{}}Conv1D [k=3, d=64, s=1] + BN1D + ReLU\\ Conv1D [k=3, d=64, s=1] + BN1D + ReLU\end{tabular} \\
\midrule
Stage3 & \begin{tabular}[c]{@{}l@{}}Conv1D [k=3, d=128, s=1] + BN1D + ReLU\\ Conv1D [k=3, d=128, s=1] + BN1D + ReLU\end{tabular} \\
\midrule
Predict & Global Average Pooling + Linear \\
\midrule
\multicolumn{2}{c}{\textbf{Architecture of the audio branch}} \\
\midrule
Stage1 & \begin{tabular}[c]{@{}l@{}}Conv1D [k=3, d=64] + BN1D + ReLU + MaxPool1d [2x1]\\ Conv1D [k=3, d=128] + BN1D + ReLU + MaxPool1d [2x1]\end{tabular} \\
\midrule
Stage2 & \begin{tabular}[c]{@{}l@{}}Conv1D [k=3, d=256] + BN1D + ReLU + MaxPool1d [k=2]\\ Conv1D [k=3, d=128] + BN1D + ReLU + MPool1D [k=2]\end{tabular} \\
\midrule
Predict & Global Average Pooling + Linear \\
\bottomrule
\end{tabular}
\end{table}
\\
\textbf{Intermediate Attention-Based Fusion:}
\\
 In this paper, we propose a fusion method based on the attention mechanism for the trifold modality data input to our model. Given the feature representations of the three modalities as $\phi_{a}$, $\phi_{v}$, and $\phi_{r}$, similar to the conventional attention mechanism, we calculate queries and keys, and learn the corresponding weight matrices within the model. For the feature representations of these three modalities, we compute the similarity matrices in pairs, with the process as follows.
\begin{equation}
A_{ij} = \text{softmax}\left(\frac{\Phi_i W_q W_k^T \Phi_j^T}{\sqrt{d}}\right), \quad \text{for } i, j \in \{a, v, r\}
\end{equation}
The softmax activation introduces a competitive mechanism within the attention mapping matrix, effectively highlighting the most critical features or temporal steps. By applying softmax normalization to each feature across the modalities, the model assigns a significance score for each key relative to each query within every modality. This means that the model can assess the importance of each feature in modality a relative to modality b. Consequently, we can aggregate the significance scores for each attribute with respect to the overall attributes of modality b, providing a composite score for each feature within modality a. The resulting attention vector is capable of accentuating prominent features in modality a, which is crucial for the model when processing and analyzing multimodal information.
\\
The fusion approach we employ does not amalgamate features of the two modalities directly. Instead, it pinpoints the most salient attributes within each modality and aligns them with similarity scores derived from the data of another modality. Thus, features that are consistent across modalities substantially contribute to the final prediction, steering the model towards learning features that are pivotal for decision-making or those that exhibit a high degree of consistency between features. This methodology facilitates the sharing of information between modalities without enforcing a strong dependency on the features learned in different branches, using only attention scores for the purpose of fusion. This strategy not only enriches the model's interpretability but also bolsters its capacity to discern nuanced patterns intrinsic to multimodal datasets.
\section{Dataset}
To explore and develop multimodal models capable of real-time assessment of driving safety, there is a need for tri-dimensional data encompassing facial videos, audio, and forward-facing vehicle camera footage during driving, with simulated and annotated scenarios representing both normal and potentially hazardous driving conditions.
Given the current absence of publicly available datasets of this nature, we have undertaken the creation of such a dataset to fill this gap. Our dataset integrates simulated driving environments with real-world driving scenarios, aimed at providing a rich data resource for the training and validation of multimodal driving behavior analysis models. We utilized the open-source project AirSimNH from Airsim and the popular video game "Need for Speed" as platforms for data generation.
\subsection{Data Generation Environment}
We employed the virtual environment provided by the AirSimNH project within AirSim, utilizing Airsim's Python interface for environmental control and data acquisition. Within this setting, road obstacles were periodically and randomly generated on the driving path to simulate sudden incidents under various driving conditions. Additionally, high-speed driving scenarios from "Need for Speed" were used to augment data diversity, particularly for simulating high-risk driving behaviors.
\\The dataset comprises three types of files: facial videos, facial audio, and forward-facing vehicle camera footage. Each file type is sampled and saved in 3-second units. Notably, scenes involving accidents or potential hazardous driving conditions are marked in the filename to indicate a dangerous driving situation
\subsection{Dataset Structure and Annotation}
 Facial Video Files (format: 01\_01\_1.mp4): These record the simulated driver's facial expressions and head movements, where the first '01' in the filename denotes the sequence number of data collection, the second field '01' or '02' represents safe and dangerous driving conditions respectively, and the final digit '1' indicates the first sample in the sequence.\\
Facial Audio Files (format: 01\_01\_1.wav): These are audio files recorded synchronously, capturing the simulated driver's vocal reactions, with a file naming structure identical to that of the facial videos.\\
Road Condition Video Files (format: 01\_01\_1\_r.mp4): Captured from the perspective of a dashcam, these provide visual information about the driving environment and conditions, with a file naming structure similar to the facial videos but with an added '\_r' suffix for differentiation.
Samples representing accidents or potential hazardous situations are marked with '02' in their filenames, indicating a dangerous driving condition. This clear annotation facilitates subsequent model training and evaluation, enabling researchers to quickly identify and utilize these critical data points.
\subsection{Data Quality and Application Potential
}
Throughout the data collection process, we have maintained strict control over the quality of the simulated environments and the authenticity of the data, ensuring the dataset accurately reflects the multimodal characteristics of various driving conditions. This dataset not only offers valuable resources for research into hazardous driving detection but also serves as an experimental foundation for developing and testing multimodal driving behavior analysis models. It is particularly suited for studies involving multimodal machine learning models, real-time monitoring of driver states and external driving environments in autonomous driving systems, and especially in detecting and preventing hazardous driving behaviors.
\section{Experiment}
In the construction of a tri-modal model for driver state monitoring, employing the same preprocessing method and feature extraction architecture for both the driver's facial video and the road condition video is justified. This approach ensures consistency in the feature space for visual information from different sources, facilitating the model's ability to integrate and interpret these inputs effectively. The unified framework enhances model stability and performance and offers convenience for future modal additions or adjustments.
\subsection{Audio Preprocessing}

The audio files underwent a preprocessing routine to ensure uniformity in length, encompassing the following steps:

\begin{enumerate}
    \item \textbf{Audio Loading:} Audio files were loaded using the \texttt{librosa.core.load} function with a standard sampling rate of 22,050 Hz, ensuring a consistent sampling rate across different audio files.
    \item \textbf{Length Normalization:} The audio files were processed to maintain a fixed length of 3.6 seconds. If the audio was shorter than 3.6 seconds, zeros were padded at the end to extend the length; if it was longer, the audio was trimmed equally from both ends to ensure consistency in length.
    \item \textbf{Audio Saving:} The modified audio was saved, keeping the original audio intact while creating a version that meets the length requirement.
\end{enumerate}

\subsection{Video Preprocessing}

The video files were also preprocessed to meet the model input requirements, involving the following steps:

\begin{enumerate}
    \item \textbf{Frame Selection and Processing:} For each video, a distributed selection of 15 frames was extracted for subsequent facial detection and cropping. This strategy aimed to uniformly sample frames from the video, regardless of its original frame count.
    \item \textbf{Facial Detection:} The MTCNN (Multi-task Cascaded Convolutional Networks) was employed to detect faces in the video frames and crop the facial regions. This step focused the model's attention on facial expressions in the videos, enhancing the accuracy of emotion recognition.
    \item \textbf{Frame Cropping and Resizing:} The detected facial regions were cropped and resized to 224x224 pixels, ensuring uniformity in all video frames to satisfy the subsequent model input requirements.
    \item \textbf{Data Saving:} The processed video frames were saved in array format, facilitating further model training and validation. Additionally, the processed frames could be encoded back into AVI format videos if required.
    \item \textbf{Processing Record:} Details of the entire processing routine, including the number of files successfully processed and a list of failed videos, were documented for further analysis and review.
\end{enumerate}
Through these steps, uniformity and standardization of the audio and video data were ensured, preparing a well-curated dataset for subsequent model training and validation. This preprocessing strategy aids in enhancing the model's performance by reducing the variability in the data that the model needs to address, allowing it to focus more on learning the core features of the data.

\subsection{Training Process}
In the experimental section of this paper, we initially utilize predefined loaders to load preprocessed driver facial video data, driver audio data, and road condition information data, and perform data augmentation on these datasets. In our experiments, we employ various masking techniques for data augmentation, enhancing the model's robustness by mixing and modifying multimodal input data (audio, visual, road conditions). This method adjusts the blend of each modal input through randomly generated coefficients and introduces all-zero data to simulate extreme noise conditions. Additionally, by combining and reordering these coefficient-weighted and unweighted input data, we generate a rich and varied set of training samples. This technique not only increases the model's tolerance to noise but also improves its adaptability in the face of data incompleteness, thereby enhancing the model's generalization capabilities.
\\
For feature extraction from road conditions and the driver's video data, we employ depthwise separable convolutions based on the EfficientFace framework to reduce model parameters and accelerate training. Residual connections are also used to speed up convergence. Moreover, self-attention mechanisms are incorporated in the feature extraction of video data to extract crucial information and efficiently capture the spatiotemporal relationships between video frames.
\\
For audio data, we utilize one-dimensional convolutional layers to capture the temporal dependencies and local features within the audio data, and employ batch normalization to accelerate the training process.
\\
Further, we apply attention mechanisms for the mutual fusion of audio, driver's facial video, and road condition video information and use the fused information for decision-making.
\\
During the training process, we use the cross-entropy loss function and the Stochastic Gradient Descent (SGD) optimizer, replacing traditional gradient descent methods with momentum to update network parameters. This helps avoid the influence of local minima and ensures the effectiveness and efficiency of the learning process. Moreover, we introduce a dynamic learning rate scheduling strategy that adaptively adjusts the learning rate based on training progress, accelerating convergence while avoiding premature local optima. To further enhance the model's generalization capability, we integrate the dropout regularization mechanism, effectively reducing the risk of overfitting.

\begin{figure}[htbp]
  \centering  \includegraphics[width=0.8\textwidth]{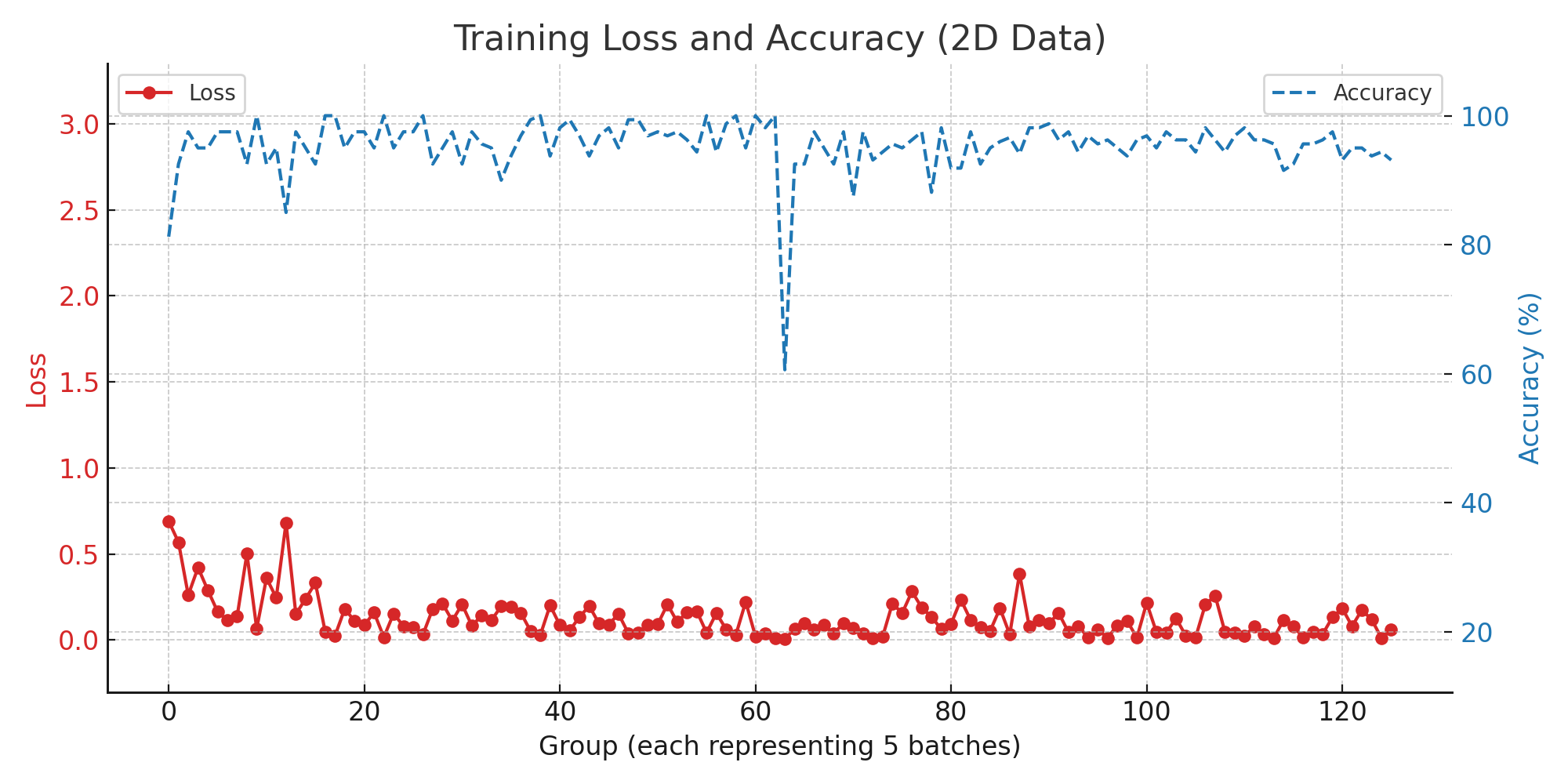}
  \caption{Traing Process}
  \label{fig:loss-accuracy}
\end{figure}

To comprehensively evaluate the overall performance of the model, we meticulously design three ablation experiments. These experiments aim to compare the specific impacts of different modal inputs (including driver facial images, audio signals, and road videos) on the final model performance and validate the effectiveness of our proposed multimodal model. Models are trained using solely single modalities (driver facial, audio, or road video), a bimodal model combining video and audio, and a complete multimodal model integrating driver facial, audio, and road video information. To ensure the effectiveness and reliability of the performance comparison, all models are trained under consistent conditions in the experimental setup. Through this systematic comparison, we are able to thoroughly evaluate the contribution of each input modality to the final model performance and how they work together to improve classification accuracy.
\begin{figure}[htbp]
  \centering  \includegraphics[width=0.8\textwidth]{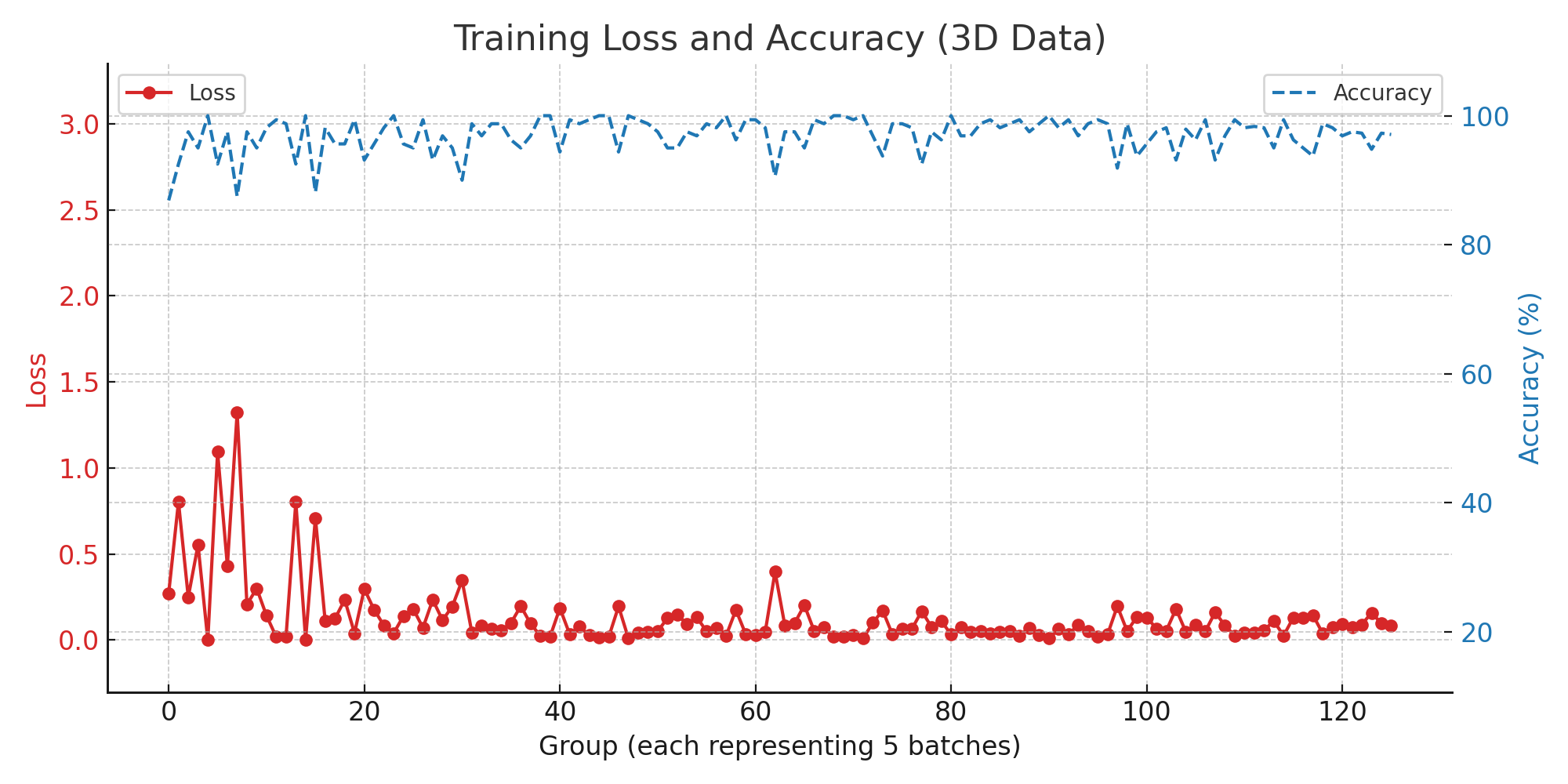}
  \caption{Traing Process}
  \label{fig:loss-accuracy}
\end{figure}
\section{Result}
Comprehensive experimental evaluations conducted on the constructed dataset indicate that the proposed three-dimensional model is capable of achieving outstanding performance in dangerous driving state recognition. The results of the comparative experiments are displayed in Table 2.

Specifically, the model attained a 96.875\% accuracy rate in the task of dangerous driving state recognition, better than the contrastive models that we set which are trained solely on any two-dimensional data (such as driver video+audio, driver video+road video, etc.) or one-dimensional data (such as driver video alone, road video alone, driver audio alone). This result underscores that use triflod-multimodal that we built can  the effectiveness and necessity of multimodal data fusion in this task.
\begin{table}[ht]
\centering
\caption{Experiment Result}
\resizebox{0.7\textwidth}{!}{
\begin{tabular}{lccccccc}
  \toprule
   & A-V-R & A-V & A-R & V-R & A & V & R \\
  \midrule
  accuracy(\%) & 96.875 & 93.75 & 81.25 & 90.625 & 46.875 & 87.5 & 78.125 \\
  loss & 0.0122 & 0.0254 & 0.4135 & 0.1911 & 0.6715 & 0.2478 & 0.5636 \\
  \bottomrule
\end{tabular}
}
\end{table}
% \begin{table}[h!]
% \centering
% \caption{Comparison of Model Performance}
% \label{tab:model_performance}
% \setlength{\tabcolsep}{14pt} % Increase the space between columns
% \renewcommand{\arraystretch}{1.5} % Increase the space between rows
% \begin{tabular}{@{}lcc@{}}
% \toprule
% Model         & Loss   & Accuracy (\%) \\ \midrule
% WideBranchNet & 0.0640 & 92.45         \\
% ResNet152     & 0.4191 & 86.12         \\
% ResNet101     & 0.4122 & 86.10         \\
% ResNet50      & 0.5146 & 82.72         \\
% Our Model     & 0.0122 & 96.875        \\ \bottomrule
% \end{tabular}
% \end{table}

Further analysis reveals that the three modalities of data play a complementary and synergistic role in recognition. The information contained within driver video data, such as facial expressions, gaze, and head posture, can reflect the driver’s level of attention and fatigue. When the driver exhibits states of fatigue or distraction, their facial expressions may become dull, their gaze unfocused, and frequent changes in head posture may occur. 

In the event of a driving accident, the driver may exhibit dramatic changes in facial expression in a short time, potentially showing focused gaze with dilated pupils, along with changes in head posture. Capturing and understanding these subtle visual cues is crucial for timely identification of dangerous driving states.

Audio data provides another important dimension. The driver’s voice, breathing sounds, and yawning noises contain key clues about their emotional state and level of alertness. For example, when a driver's emotions are heightened and their speech rate increases, it often indicates a state of tension or anxiety; conversely, slower breathing and yawning sounds suggest the risk of fatigued driving. Through voice emotion recognition and acoustic event detection technologies, we can extract these critical pieces of information from audio data to aid in the judgment of dangerous driving states.

Road video data, from the perspective of the external environment, provides crucial information about the driving scene in which the vehicle is situated. Through the analysis of road video, we can gather a series of indicators such as the vehicle’s speed, the distance to the vehicle ahead, and lane deviation. These indicators collectively reflect the driver's behavior and the potential risks they face. For instance, frequent and prolonged lane deviations suggest that the driver is not concentrating and may engage in dangerous driving.

The fusion of driver video, audio, and road video modalities enables a comprehensive portrayal and understanding of driving risks from multiple perspectives, including the driver’s state, driving behavior, and driving environment. The combination of these three enhances the model's capability for scene understanding and state judgment. The complementary and synergistic information within multimodal data significantly improves the accuracy, reliability, and robustness of dangerous driving state recognition, providing a more comprehensive and dependable perception capability for intelligent driving safety assistance systems.

To compare our model with current video anomaly detection models, we replicated several anomaly detection models, including AstNet\cite{le2023attention} and WideBranchNet, and tested them on our custom dataset. We used our own road condition video data to evaluate these models. Since different anomaly detection models employ varying input architectures, we uniformly utilized a data loading method tailored to our simulated driving dataset. Table 6.2 displays the performance of these models and our three-dimensional model during testing.
% \begin{table}[h!]
% \centering
% \caption{Comparison of Model Performance}

% \label{tab:model_performance}
% \begin{tabular}{@{}lcc@{}}
% \toprule
% Model         & Loss   & Accuracy (\%) \\ \midrule
% WideBranchNet & 0.0640 & 92.45         \\
% ASTnet(ResNet152)     & 0.4191 & 86.12         \\
% ASTnet(ResNet101)     & 0.4122 & 86.10         \\
% ASTnet(ResNet50)      & 0.5146 & 82.72         \\
% Our Model     & 0.0122 & 96.875        \\ \bottomrule
% \end{tabular}
% \end{table}
\begin{table}[h!]
\centering
\caption{Comparison of Model Performance}
\resizebox{0.5\textwidth}{!}{ % Adjust this value to change the scale of the table
\begin{tabular}{@{}lcc@{}}
\toprule
Model         & Loss   & Accuracy (\%) \\ \midrule
WideBranchNet & 0.0640 & 92.45         \\
ASTnet(ResNet152)     & 0.4191 & 86.12         \\
ASTnet(ResNet101)     & 0.4122 & 86.10         \\
ASTnet(ResNet50)      & 0.5146 & 82.72         \\
Our Model     & 0.0122 & 96.875        \\ \bottomrule
\end{tabular}
}
\end{table}

The training processes of two-dimensional and three-dimensional models, as demonstrated in Figures 1 and 2, show that although both models experienced initial fluctuations, they subsequently converged rapidly to higher levels of accuracy. However, compared to the two-dimensional model, the three-dimensional model exhibited greater stability throughout the training process, and its final training accuracy was generally higher. This suggests that due to the higher complexity of the three-dimensional model, it may possess superior generalization capabilities. During the processing of three-dimensional data, the three-dimensional model is able to learn more features and patterns, thus displaying stronger performance in both training and prediction phases.
\newpage

\section{Conclusion}
We have developed a road incident recognition model that harnesses three distinct data sources: driver video, audio data, and road condition videos. By utilizing an attention-based intermediate layer fusion technique, we have efficiently integrated these three types of data, thus constructing an end-to-end multimodal recognition system. We constructed a simulated driving dataset that includes tridimensional data and tested our proposed multimodal model on it. The test results demonstrate that our trifold modality model significantly outperforms the unimodal and bimodal models in performance, underscoring its potential value in practical applications.

The three modalities of data, namely driver video, audio, and road video, play a complementary and synergistic role in recognition. The information contained in driver video, such as facial expressions, gaze, and head posture, can reflect the driver's level of attention and fatigue. The driver's voice, breathing sounds, and yawning noises in audio data contain crucial clues about their emotional state and level of alertness. Road video data, from the perspective of the external environment, provides key information about the driving scene in which the vehicle is situated. The fusion of these three modalities enables the model to comprehensively portray and understand driving risks from multiple angles, including the driver's state, driving behavior, and driving environment, significantly enhancing the accuracy, reliability, and robustness of dangerous driving state recognition.

Comparative experiments with current video anomaly detection models demonstrate that the proposed three-dimensional model outperforms models such as AstNet and WideBranchNet on the custom-built dataset. This further confirms the advantages of multimodal fusion and three-dimensional modeling approaches in this task.

The comparison of training processes between two-dimensional and three-dimensional models shows that although both models experience fluctuations in the initial training stage, they quickly converge to high levels of accuracy. However, compared to the two-dimensional model, the three-dimensional model exhibits greater stability throughout the training process, and its final training accuracy is generally higher. This suggests that due to the higher complexity of the three-dimensional model, it may possess stronger generalization capabilities. During the processing of three-dimensional data, the three-dimensional model can learn more features and patterns, thus demonstrating superior performance in both training and prediction phases.

The three-dimensional multimodal driving anomaly detection model has demonstrated exceptional performance and has broad prospects for practical applications. In the future, this model can be further improved and expanded in the following aspects:
\begin{itemize}
\item Expanding the dataset: The current model's dataset was collected during our simulated driving process using a driving simulator. To improve the model's robustness and generalization ability, we can collect more real-world driving data, covering a wider range of driving scenarios, driving states, and anomalous situations. This will help the model achieve better performance in practical applications.
\item Introducing more modal data: In addition to driver video, audio, and road video data, we can also introduce other types of sensor data, such as vehicle speed, acceleration, or distance information of road obstacles. These additional data can provide the model with more comprehensive driving state information, which helps to further improve the accuracy of anomaly detection.
\item Optimizing the model structure: We can explore more advanced attention mechanisms and fusion strategies to more effectively integrate information from different modalities.
\end{itemize}
\newpage

\newpage
\bibliographystyle{unsrt}
\bibliography{stability}
	
\end{document}